%% file: main.tex
\documentclass[runningheads]{llncs}
\usepackage[T1]{fontenc}

\usepackage{graphicx}
\usepackage{xcolor}
\input{pre}

\begin{document}

\title{VesselVAE: Recursive Variational Autoencoders for 3D Blood Vessel Synthesis}

\titlerunning{Recursive Variational Autoencoders for 3D Blood Vessel Synthesis}

\author{Paula Feldman \inst{1,3} 
\and Miguel Fainstein \inst{3} 
\and Viviana Siless \inst{3} 
\and Claudio Delrieux\inst{1,2} 
\and Emmanuel Iarussi\inst{1,3}} 

\authorrunning{Paula Feldman et al.}

\institute{ Consejo Nacional de Investigaciones Científicas y Técnicas, Argentina \and{Universidad Nacional del Sur, Bahía Blanca, Argentina} \and{Universidad Torcuato Di Tella, Buenos Aires, Argentina} \\
\email{paulafeldman@conicet.gov.ar}}

\maketitle            

\begin{abstract}
We present a data-driven generative framework for synthesizing blood vessel 3D geometry. 
This is a challenging task due to the complexity of vascular systems, which are highly variating in shape, size, and structure.
Existing model-based methods provide some degree of control and variation in the structures produced, but fail to capture the diversity of actual anatomical data.
We developed VesselVAE, a recursive variational Neural Network that fully exploits the hierarchical organization of the vessel and learns a low-dimensional manifold encoding branch connectivity along with geometry features describing the target surface. 
After training, the VesselVAE latent space can be sampled to generate new vessel geometries. 
To the best of our knowledge, this work is the first to utilize this technique for synthesizing blood vessels. 
We achieve similarities of synthetic and real data for radius (.97), length (.95), and tortuosity (.96).
By leveraging the power of deep neural networks, we generate 3D models of blood vessels that are both accurate and diverse, which is crucial for medical and surgical training, hemodynamic simulations, and many other purposes.

\keywords{Vascular 3D model  \and Generative modeling \and Neural Networks.}
\end{abstract}

\section{Introduction}
Accurate 3D models of blood vessels are increasingly required for several purposes in Medicine and Science~\cite{talou2021adaptive}. 
These meshes are typically generated using either image segmentation or synthetic methods.
Despite significant advances in vessel segmentation~\cite{tetteh2020deepvesselnet}, reconstructing thin features accurately from medical images remains challenging~\cite{alblas2021deep}.
Manual editing of vessel geometry is a tedious and error prone task that requires expert medical knowledge, which explains the scarcity of curated datasets. 
As a result, several methods have been developed to adequately synthesize blood vessel geometry~\cite{WU20134}.

Within the existing literature on generating vascular 3D models, we identified two primary types of algorithms: fractal-based, and space-filling algorithms.
Fractal-based algorithms use a set of fixed rules that include different branching parameters, such as the ratio of asymmetry in arterial bifurcations and the relationship between the diameter of the vessel and the flow ~\cite{galarreta2013three,zamir2001arterial}. 
On the other hand, space-filling algorithms allow the blood vessels to grow into a specific perfusion volume while aligning with hemodynamic laws and constraints on the formation of blood vessels~\cite{hamarneh2010vascusynth,talou2021adaptive,schneider2012tissue,merrem2017computational,rauch2021interactive}. 
Although these \emph{model-based} methods provide some degree of control and variation in the structures produced, they often fail to capture the diversity of real anatomical data.

In recent years, deep neural networks led to the development of powerful generative models~\cite{xu2023generative}, such as Generative Adversarial Networks~\cite{goodfellow2020generative,kazeminia2020gans} and Diffusion Models~\cite{ho2020denoising}, which produced groundbreaking performance in many applications, ranging from image and video synthesis to molecular design. 
These advances have inspired the creation of novel network architectures to model 3D shapes using voxel representations~\cite{wu2016learning}, point clouds~\cite{yang2019pointflow}, signed distance functions~\cite{park2019deepsdf}, and polygonal meshes~\cite{nash2020polygen}.
In particular, and close to our aim, Wolterink~et~al.~\cite{wolterink2018blood} propose a GAN model capable of generating coronary artery anatomies. 
However, this model is limited to generating single-channel blood vessels and thus does not support the generation of more complex, tree-like vessel topologies. 

In this work we propose a novel \emph{data-driven} framework named VesselVAE for synthesizing blood vessel geometry.
Our generative framework is based on a Recursive variational Neural Network (RvNN), that has been applied in various contexts, including natural language~\cite{socher2011parsing,socher2014recursive}, shape semantics modeling~\cite{li2017grass,li2019grains}, and document layout generation~\cite{patil2020read}. 
In contrast to previous data-driven methods, our recursive network fully exploits the hierarchical organization of the vessel and learns a low-dimensional manifold encoding branch connectivity along with geometry features describing the target surface. 
Once trained, the VesselVAE latent space is sampled to generate new vessel geometries. 
To the best of our knowledge, this work is the first to synthesize multi-branch blood vessel trees by learning from real data.
Experiments show that synth and real blood vessel geometries are highly similar: radius (.97), length (.95), and tortuosity (.96).

\section{Methods}
\begin{figure}[t]
\includegraphics[width=\textwidth]{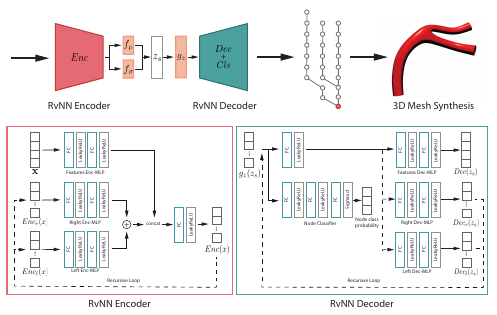}
\caption{Top: Overview of the Recursive variational Neural Network for synthesizing blood vessel structures. 
The architecture follows an Encoder-Decoder framework which can handle the hierarchical tree representation of the vessels. 
VesselVAE learns to generate the topology and attributes for each node in the tree, which is then used to synthesize 3D meshes. 
Bottom: Layers of the Encoder and Decoder networks comprising branches of fully-connected layers followed by leaky ReLU activations. 
Notice that right/left Enc-MLPs of the Encoder only execute when the incoming tree requires it. 
Similarly, the Decoder only uses right/left Dec-MLPs when the Node Classifier predicts bifurcations.}
\label{fig:overview}
\end{figure}

\textbf{Input.} The network input is a binary tree representation of the blood vessel 3D geometry. 
Formally, each tree is defined as a tuple $(T, \mathcal{E})$, where $T$ is the set of nodes, and $\mathcal{E}$ is the set of directed edges connecting a pair of nodes ${(n, m)}$, with $ n,m \in T$.
In order to encode a 3D model into this representation, vessel segments $V$ are parameterized by a central axis consisting of ordered points in Euclidean space: $V = {v_1, v_2, \ldots, v_N}$ and a radius $r$, assuming a piece-wise tubular vessel for simplicity. 
We then construct the binary tree as a set of nodes $ T = {n_1, n_2, \ldots, n_N}$, where each node $n_i$ represents a vessel segment $v$ and contains an attribute vector 
$\mathbf{x}_i = [x_i, y_i, z_i, r_i] \in \mathbb{R}^4$ 
with the coordinates of the corresponding point and its radius $r_i$. See Section~\ref{sec:experimental} for details. \\

\textbf{Network architecture.} 
The proposed generative model is a Recursive variational Neural Network (RvNN) consisting of two main components: the Encoder ($Enc$) and the Decoder ($Dec$) networks. 
The role of the Encoder is to transform a tree structure into a hierarchical encoding situated on the learned manifold. 
On the other hand, the Decoder network is capable of sampling from this encoded space utilizing these samples to decode tree structures, as depicted in Fig. ~\ref{fig:overview}.
The encoding and decoding processes are achieved through a depth-first traversal of the tree, where each node is combined with its parent node in a recursive manner.
The model outputs a hierarchy of vessel branches, where each internal node in the hierarchy is represented by a vector encoding its own attributes and the information of all subsequent nodes in the tree.

Within the RvNN Decoder network there are two essential components: the Node Classifier ($Cls$) and the Features Decoder Multi-Layer Perceptron (Features Dec-MLP). The Node Classifier discerns the type an encoded node should be decoded into, whether a leaf node or an internal node with one or two bifurcations. 
This is implemented as a multi-layer perceptron trained to predict a three-category bifurcation probability with an encoded vector as input. Complementing the Node Classifier, the Features Dec-MLP is responsible for reconstructing the attributes of each node, specifically its position and radius. Furthermore, two additional components, the Right and Left Dec-MLP, are in charge of recursively decoding the next encoded vector in the tree hierarchy. These decoder branches do not always execute, and depend on the decision made by the classifier. If the Node Classifier predicts one bifurcation for a node, a right child is assumed by default. 

In addition to the core architecture, our model is further augmented with three auxiliary, shallow, fully-connected neural networks: $f_\mu$, $f_\sigma$, and $g_z$. Positioned before the RvNN bottleneck, the $f_\mu$ and $f_\sigma$ networks shape the distribution of the latent space where encoded tree structures lie. Conversely, the $g_z$ network, situated after the bottleneck, facilitates the decoding of latent variables, aiding the Decoder network in the reconstruction of tree structures. Collectively, these supplementary networks streamline the data transformation process through the model. All activation functions used in the network are leaky ReLUs. See the Appendix for other implementation details.\\

\textbf{Objective.} Our generative model is trained to learn a probability distribution over the latent space that can be used to generate new blood vessel segments.
After encoding, the decoder takes samples from a multivariate Gaussian distribution: $z_s(x) \sim N(\mu, \sigma)$ with $\mu=f_\mu(Enc(x))$ and $\sigma=f_\sigma(Enc(x))$, where $Enc$ is the recursive encoder and $f_\mu,f_\sigma$ are two fully-connected neural networks.
In order to recover the feature vectors $\mathbf{x}$ for each node along with the tree topology, we simultaneously train the regression network (Features Dec-MLP in Fig. \ref{fig:overview}) on a reconstruction objective $L_{recon}$, and the Node Classifier using $L_{topo}$.
Additionally, in line with the general framework proposed by $\beta$-VAE~\cite{higgins2017beta}, we incorporated a Kullback-Leibler (KL) divergence term encouraging the distribution $p(z_s(x))$ over all training samples $x$ to move closer to the prior of the standard normal distribution $p(z)$.
We therefore minimize the following equation: 
\begin{equation}
    L = L_{recon} + \alpha L_{topo} + \gamma L_{KL},
\label{eq:loss}
\end{equation}
where the reconstruction loss is defined as $L_{recon} = \left\|Dec\left(z_s(x)\right)-x\right\|_2 $, the Kullback-Leibler divergence loss is $L_{\mathrm{KL}}=D_{\mathrm{KL}}\left(p\left(z_s(x)\right) \| p(z)\right)$, and the topology objective is a three-class cross entropy loss $L_{topo} =\Sigma_{c=1}^{3} x_c \log(Cls(Dec(x))_c)$.
Notice that $x_{c}$ is a binary indicator (0 or 1) for the true class of the sample $x$. 
Specifically, $x_{c} = 1$ if the sample belongs to class $c$ and 0 otherwise.
$Cls(Dec(x))_c$ is the predicted probability of the sample $x$ belonging to class $c$ (zero, one, or two bifurcations), as output by the classifier. 
Here, $Dec(x)$ denotes the encoded-decoded node representation of the input sample $x$.\\

\textbf{3D mesh synthesis.} Several algorithms have been proposed in the literature to generate a surface 3D mesh from a tree-structured centerline~\cite{WU20134}. 
For simplicity and efficiency, we chose the approach described in~\cite{felkel2004surface}, which produces good quality meshes from centerlines with a low sample rate.
The implemented method iterates through the points in the curve generating a coarse quadrilateral mesh along the segments and joints. 
The centerline sampling step is crucial for a successful reconstruction outcome.
Thus, our re-sampling is not equispaced but rather changes with curvature and radius along the centerline, increasing the frequency of sampling near high-curvature regions.
This results in a better quality and more accurate mesh.
Finally, Catmull-Clark subdivision algorithm~\cite{CATMULL1978350} is used to increase mesh resolution and smooth out the surface.  

\section{Experimental Setup} \label{sec:experimental}

\begin{figure}[t]
\includegraphics[width=\textwidth]{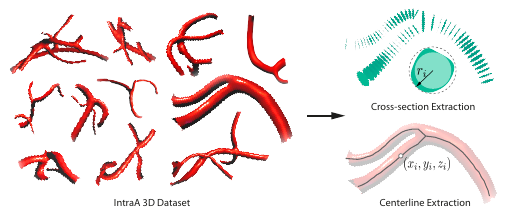}
\caption{Dataset and pre-processing overview: The raw meshes from the IntraA 3D collection undergo pre-processing using the VMTK toolkit. This step is crucial for extracting centerlines and cross-sections from the meshes, which are then used to construct their binary tree representations.}
\label{fig:dataset}
\end{figure}

\textbf{Materials.} We trained our networks using a subset of the open-access IntrA dataset~\footnote{\url{https://github.com/intra3d2019/IntrA}} published by Yang~et~al.~in 2020~\cite{yang2020intra}. 
This subset consisted of 1694 healthy vessel segments reconstructed from 2D MRA images of patients. 
We converted 3D meshes into a binary tree representation and used the \emph{network extraction} script from the VMTK toolkit~\footnote{\url{http://www.vmtk.org/vmtkscripts/vmtknetworkextraction}} to extract the centerline coordinates of each vessel model.
The centerline points were determined based on the ratio between the sphere step and the local maximum radius, which was computed using the advancement ratio specified by the user. 
The radius of the blood vessel conduit at each centerline sample was determined using the computed cross-sections assuming a maximal circular shape (See Figure~\ref{fig:dataset}).
To improve computational efficiency during recursive tree traversal, we implemented an algorithm that balances each tree by identifying a new root.
We additionally trimmed trees to a depth of ten in our experiments. 
This decision reflects a balance between the computational demands of depth-first tree traversal in each training step and the complexity of the training meshes. 
Trees with higher depth and non-binary bifurcations or loops were excluded from our study. 
However, non-binary trees can be converted into binary trees and it is possible to train with deeper trees at the expense of higher computational costs.
Ultimately, we were able to obtain 700 binary trees from the original meshes using this approach.\\

\textbf{Implementation details.} For the centerline extraction, we set the advancement ratio in the VMTK script to $1.05$. 
The script can sometimes produce multiple cross-sections at centerline bifurcations. 
In those cases, we selected the sample with the lowest radius, which ensures proper alignment with the centerline principal direction. 
All attributes were normalized to a range of $[0, 1]$.
For the mesh reconstruction we used 4 iterations of Catmull-Clark subdivision algorithm. 
The data pre-processing pipeline and network code were implemented in Python and PyTorch Framework.\\

\textbf{Training.} In all stages, we set the batch size to 10 and used the ADAM optimizer with $\beta_1$ = 0.9, $\beta_2$ = 0.999, and a learning rate of $1\times10^{-4}$. 
We set $\alpha=.3$ and $\gamma=.001$ for Equation~\ref{eq:loss} in our experiments.
To enhance computation speed, we implemented dynamic batching~\cite{looks2017deep}, which groups together operations involving input trees of dissimilar shapes and different nodes within a single input graph. 
It takes approximately 12 hours to train our models on a workstation equipped with an NVIDIA A100 GPU, 80GB VRAM, and 256GB RAM.
However, the memory footprint during training is very small ($\leq$1GB) due to the use of a lightweight tree representation. 
This means that the amount of memory required to store and manipulate our training data structures is minimal. 
During training, we ensure that the reconstructed tree structure aligns with the original tree's structure, rather than relying solely on the classifier's predictions. 
We train the classifier using a cross-entropy loss that compares its predictions to the actual values from the original tree. 
Since the number of nodes in each class is unbalanced, we scale the weight given to each class in the cross-entropy loss using the inverse of each class count. 
During preliminary experiments, we observed that accurately classifying nodes closer to the tree root is critical. 
This is because a miss-classification of top nodes has a cascading effect on all subsequent nodes in the tree (i.e. skip reconstructing a branch). 
To account for this, we introduce a weighting scheme that assigns a weight to each node's cross-entropy loss based on the number of total child nodes.
This weight is then normalized by the total number of nodes in the tree.\\

\textbf{Metrics.} We defined a set of metrics to evaluate our trained network's performance.  
By using these metrics, we can determine how well the generated 3D models of blood vessels match the original dataset distribution, as well as the diversity of the generated output. 
The chosen metrics have been widely used in the field of blood vessel 3D modeling, and have shown to provide reliable and accurate quantification of blood vessels main characteristics~\cite{lang2012three,bullitt2003vascular}.
We analyzed the tortuosity per branch, the vessel centerline total length, and the average radius of the tree.
Tortuosity distance metric~\cite{bullitt2003measuring} is a widely used metric in the field of blood vessel analysis (mainly because of its clinical importance).
It measures the amount of twistiness in each branch of the vessel.
The vessel total length and average radius were used in previous work to distinguish healthy vasculature from cancerous malformations.
Finally, in order to measure the distance across distributions for each metric, we compute the cosine similarity.

\section{Results}
\begin{figure}[t]
\includegraphics[width=\textwidth]{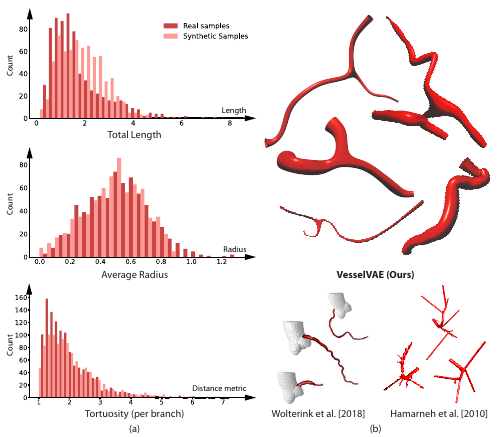}
\caption{(a) shows the histograms of total length, average radius and tortuosity per branch for both, real and synthetic samples. (b) shows a visual comparison among our method and two baselines~\cite{wolterink2018blood,hamarneh2010vascusynth}.}
\label{fig:results}
\end{figure}

We conducted both quantitative and qualitative analyses to evaluate the model's performance. For the quantitative analyses, we implemented a set of metrics commonly used for characterizing blood vessels. We computed histograms of the radius, total length, and tortuosity for the real blood vessel set and the generated set (700 samples) in Figure~\ref{fig:results} (a).  The distributions are aligned and consistent. We measured the closeness of histograms with the cosine similarity by projecting the distribution into a vector of $n$-dimensional space ($n$ is the number of bins in the histogram). Since our points are positive, the results range from 0 to 1. We obtain a radius cosine similarity of .97, a total length cosine similarity of .95, and a tortuosity cosine similarity of .96. Results show high similarities between histograms demonstrating that generated blood vessels are realistic. Given the differences with the baselines generated topologies, for a fair comparison, we limited our evaluation to a visual inspection of the meshes.   

The qualitative analyses consisted of a visual evaluation of the reconstructed outputs provided by the decoder network. We visually compared them to state-of-the-art methods in Figure~\ref{fig:results} (b). The method described in Wolterink et al. [2018] \cite{wolterink2018blood} is able to generate realistic blood vessels but without branches and the method described in Hamarneh et al. [2010] \cite{hamarneh2010vascusynth} is capable of generating branches with straight shapes, missing on realistic modeling. In contrast, our method is capable of generating realistic blood vessels containing branches, with smooth varying radius, lengths, and tortuosity. 

\section{Conclusions}
 We have presented a novel approach for synthesizing blood vessel models using a variational recursive autoencoder. 
 Our method enables efficient encoding and decoding of binary tree structures and produces high-quality synthesized models. 
 In the future, we aim to explore combinations of our approach with representing surfaces by the zero level set in a differentiable implicit neural representation (INR) \cite{alblas2023going}. 
 This could lead to more accurate and efficient modeling of blood vessels and potentially other non-tree-like structures such as capillary networks. 
 Since the presented framework would require significant adaptations to accommodate such complex topologies, exploring this problem would certainly be an interesting direction for future research.
 Additionally, the generated geometries might show self-intersections. 
 In the future, we would like to incorporate restrictions into the generative model to avoid such artifacts.  
 Overall, we believe that our proposed approach holds great promise for advancing 3D blood vessel geometry synthesis and contributing to the development of new clinical tools for healthcare professionals.

\subsubsection{Acknowledgements} This project was supported by grants from Salesforce, USA (Einstein AI 2020), National Scientific and Technical Research Council (CONICET), Argentina (PIP 2021-2023 GI - 11220200102981CO), and Universidad Torcuato Di Tella, Argentina.

\bibliographystyle{splncs04}
\newpage
\bibliography{bibliography}
\newpage
\input{supplemental}
\end{document}

%% file: pre.tex
\usepackage{dsfont}
\usepackage{amsthm}
\usepackage{overpic}
\usepackage{contour}
\contourlength{1pt}
\usepackage{xspace}
\usepackage{sidecap}
\usepackage{rotating}
\usepackage[normalem]{ulem}
\usepackage{amsmath}
\usepackage{xcolor,colortbl}
\usepackage{tabularx}
\usepackage{soul}
\usepackage{url}
\usepackage{amsfonts}

\usepackage{etoolbox}

\newlength{\indentlaenge}


%% file: supplemental.tex
\title{Supplementary Materials: 
VesselVAE: Recursive Variational Autoencoders for 3D Blood Vessel Synthesis}

\titlerunning{VesselVAE}

\author{Paula Feldman \inst{1,2} 
\and Miguel Fainstein \inst{2} 
\and Viviana Siless \inst{2} 
\and Claudio Delrieux\inst{1,3} 
\and Emmanuel Iarussi\inst{1,2}}

\authorrunning{Paula Feldman}

\institute{ Consejo Nacional de Investigaciones Científicas y Técnicas, Argentina \and{Universidad Torcuato Di Tella, Buenos Aires, Argentina} 
\and{Universidad Nacional del Sur, Bahía Blanca, Argentina} \\
\email{paulafeldman@conicet.gov.ar}}

\maketitle              
\begin{table}
\centering
\label{tab:architecture}
 
\caption{Our proposed method consists of two primary components: a recursive Encoder and a recursive Decoder. All layers in each component are fully connected with leaky rectified linear unit (Leaky ReLU) activations. The Classifier is a sub-network inside the Decoder that predicts a label for each node in the input graph.}

\begin{tabular}{|l|l|l|l|l|l|}
\hline
Layer name & Input shape & Output shape & Layer name & Input shape & Output shape   \\\hline
\multicolumn{3}{|c|}{\textbf{Encoder}} & \multicolumn{3}{|c|}{\textbf{Decoder}} \\
\hline
fc1 & 4 & 512 & fc1 & 128 & 256 \\
fc2 & 512 & 128 & fc\_left1 & 256 & 256 \\
right\_fc1 & 128 & 512 & fc\_left2 & 256 & 128 \\
right\_fc2 & 512 & 128 & fc\_right1 & 256 & 256 \\
left\_fc1 & 128 & 512 & fc\_right2 & 256 & 128 \\
left\_fc2 & 512 & 128 & fc2 & 256 & 128 \\
fc3 & 256 & 128 & fc3 & 128 & 4 \\
\hline
\multicolumn{3}{|c|}{Sample Encoder ($f$)} & \multicolumn{3}{|c|}{Sample Decoder ($g$)} \\ \hline
fc1 & 128 & 512 & fc1 & 128 & 256 \\ 
fc2mu & 512 & 128 & fc2 & 256 & 128 \\ 
fc2var & 512 & 128 & - &  - & - \\ \hline
\multicolumn{3}{|c|}{\;} & \multicolumn{3}{|c|}{Classifier} \\ \hline
- & - & - & fc1 & 128 & 256 \\ 
- & - & - & fc2 & 256 & 256 \\ 
- & - & - & fc3 & 256 & 3 \\ 
\hline
\end{tabular}

\end{table}

\begin{figure}[t]
\includegraphics[width=\textwidth]{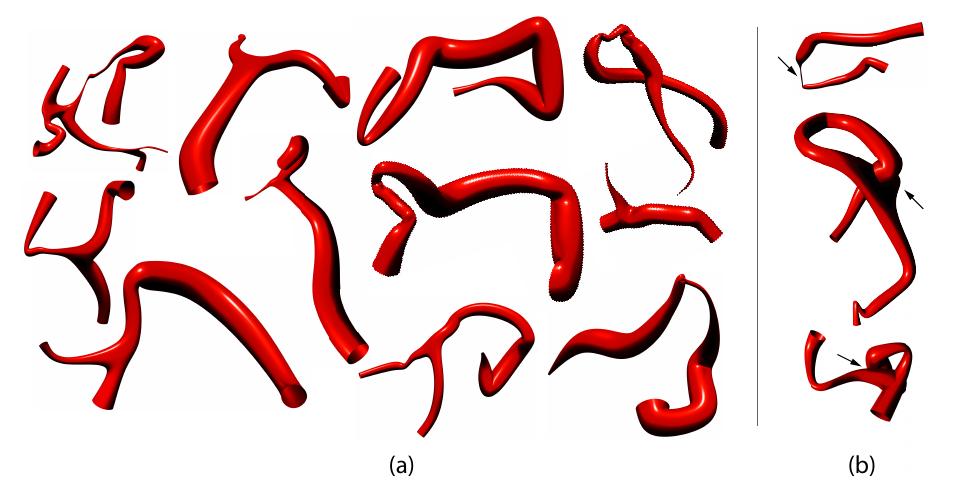}
\caption{(a) Additional renders of blood vessels generated using VesselVAE. Our approach is capable of generating diverse and intricate vessel structures, including variations in thickness, branching patterns, and curvatures. These outputs closely resemble real anatomical structures and demonstrate the effectiveness of our neural network architecture and training procedures. (b) Limitations of our method: While VesselVAE is able to generate a wide variety of complex structures, it may occasionally struggle to reproduce realistic data. For example, the sample at the top of the figure features an extremely thin segment that may not occur in real blood vessels. Additionally, the mesh reconstruction algorithm employed by our method can sometimes produce vessels with self-intersections, which are not physically plausible in biological systems.}
\label{fig:suplementario}
\end{figure}
